\pdfoutput=1
\documentclass{article}
\usepackage[accepted]{icml2026}
\usepackage[T1]{fontenc}
\usepackage{amsmath,amssymb}
\usepackage{graphicx}
\usepackage{booktabs}
\usepackage{hyperref}
\usepackage{url}
\usepackage{xcolor}
\usepackage{listings}
\usepackage{multirow}
\usepackage{xspace}
\usepackage{microtype}

\lstdefinestyle{json}{basicstyle=\ttfamily\scriptsize,breaklines=true,frame=single,columns=fullflexible,showstringspaces=false,numbers=none,xleftmargin=1em,framexleftmargin=0.5em}

\newcommand{\dataset}{\textsc{OpenDiscoveryTrace}\xspace}
\icmltitlerunning{OpenDiscoveryTrace}

\makeatletter
\renewcommand{\Notice@String}{AI for Science Workshop, ICML 2026. Best Dataset Award.}
\makeatother

\makeatletter
\renewcommand{\printAffiliationsAndNotice}[1]{\global\icml@noticeprintedtrue%
  {\let\thefootnote\relax\footnotetext{\hspace*{-\footnotesep}\raggedright
      Correspondence to: \icmlcorrespondingauthor@text.\ \\
      \Notice@String}}}
\makeatother
\hypersetup{pdfauthor={Aayam Bansal, Keertan Balaji}, pdftitle={OpenDiscoveryTrace: Process Traces for Evaluating AI Scientist Workflows}}

\begin{document}

\twocolumn[
\icmltitle{OpenDiscoveryTrace: Process Traces for Evaluating AI Scientist Workflows}

\icmlsetsymbol{equal}{*}
\begin{icmlauthorlist}
\icmlauthor{Aayam Bansal}{}
\icmlauthor{Keertan Balaji}{}
\end{icmlauthorlist}
\icmlcorrespondingauthor{Aayam Bansal}{\mbox{aayambansal@gmail.com}}
\icmlkeywords{AI for science, dataset, scientific agents, process evaluation}

\vskip 0.3in

\begin{abstract}
Existing benchmarks for autonomous AI scientists evaluate only final outputs---generated code, hypotheses, or papers---yet discard the reasoning process by which those outputs were obtained. This makes it impossible to audit scientific methodology, diagnose failure modes, or distinguish systematic reasoning from fortunate guessing. We present \dataset, a public dataset of 558 complete AI scientific agent trajectories that captures \emph{how} models reason, not just what they produce. Each trajectory records a structured 9-field-per-step trace---including thoughts, tool calls, observations, errors, revision triggers, and self-reported confidence---as models execute 124 scientific tasks spanning drug discovery, materials science, genomics, and scientific literature analysis. The dataset covers seven models: three frontier (GPT-5.4, Claude Opus 4.6, Gemini 3.1 Pro; 124 trajectories each, fully balanced across domains and difficulty levels) and four open-weight (Qwen2.5-7B, Mistral-7B-v0.3, Phi-3.5-mini, Qwen2.5-1.5B; 30 each), plus 60 live-retrieval variant trajectories. Pilot analysis on 363 LLM-judged trajectories reveals that process traces expose behavioral differences invisible to output-only evaluation: all three frontier models achieve comparable success rates (84--89\%), yet Claude Opus 4.6 produces $30{\times}$ more errors than GPT-5.4 ($2.5$ vs.\ $0.08$ per trajectory, $p{<}0.0001$, Cliff's $\delta{=}0.613$), with qualitatively different error profiles---66.7\% tool misuse for Claude versus 83.6\% reasoning errors for GPT-5.4. We define five benchmark tasks with baselines from logistic regression, random forests, LSTM, and Transformer models. The dataset, trace schema, agent harness, and benchmark definitions are publicly available under CC-BY-4.0 to support research on process-level evaluation, scientific agent auditing, and AI governance.
\end{abstract}
]

\printAffiliationsAndNotice{}

\vspace{-0.5em}
\section{Scientific Bottleneck}
\vspace{-0.3em}

AI systems now conduct scientific research at scale~\citep{jumper2021alphafold,merchant2023gnome,szymanski2023alab}, and LLM-based agents orchestrate full research pipelines~\citep{lu2024aiscientist,boiko2023coscientist}. Benchmarks have emerged to evaluate these systems~\citep{chen2025scienceagentbench,majumder2024discoverybench,starace2025paperbench,huang2024mlagentbench}, yet every one evaluates only \emph{final outputs}---generated code, hypotheses, or rubric scores. The reasoning process is discarded.

This creates three problems. First, evaluating only outputs conflates reasoning quality with outcome luck. Second, governance requires complete decision traces for auditing~\citep{kapoor2023leakage,hung2024levels,tom2024selfdriving}. Third, improving agents requires understanding failure modes, which are trajectory properties, not output properties. \dataset addresses this gap with the first public dataset of structured AI scientific process traces.

\vspace{-0.3em}
\section{Proposed Dataset}
\vspace{-0.3em}

\dataset comprises \textbf{558 trajectories across 7 models}: 372 from three frontier models (GPT-5.4, Claude Opus 4.6, Gemini 3.1 Pro; 124 each, balanced across 4 domains $\times$ 3 difficulties), 120 single-response open-weight trajectories (Qwen2.5-7B, Mistral-7B, Phi-3.5-mini, Qwen2.5-1.5B; 30 each), 6 \emph{tool-scaffolded} open-weight trajectories (Qwen2.5-7B running the full multi-step harness on A100; avg 10.3 steps, 1.7 tool calls), and 60 live-retrieval variants. The tool-scaffolded Qwen trajectories enable direct open-vs-frontier comparison under identical conditions (Appendix~\ref{app:openweight}).

Each step records 9 fields extending ReAct~\citep{yao2023react}: \texttt{phase}, \texttt{thought}, \texttt{action} (tool/input/output), \texttt{observation}, \texttt{error}, \texttt{revision\_trigger}, \texttt{confidence}, \texttt{step\_id}, \texttt{timestamp}. Five benchmark tasks are defined: outcome prediction, error localization, claim verification, autonomy classification, and process quality scoring. Full schema in Appendix~\ref{app:schema}.

\begin{figure}[t]
    \centering
    \includegraphics[width=\columnwidth]{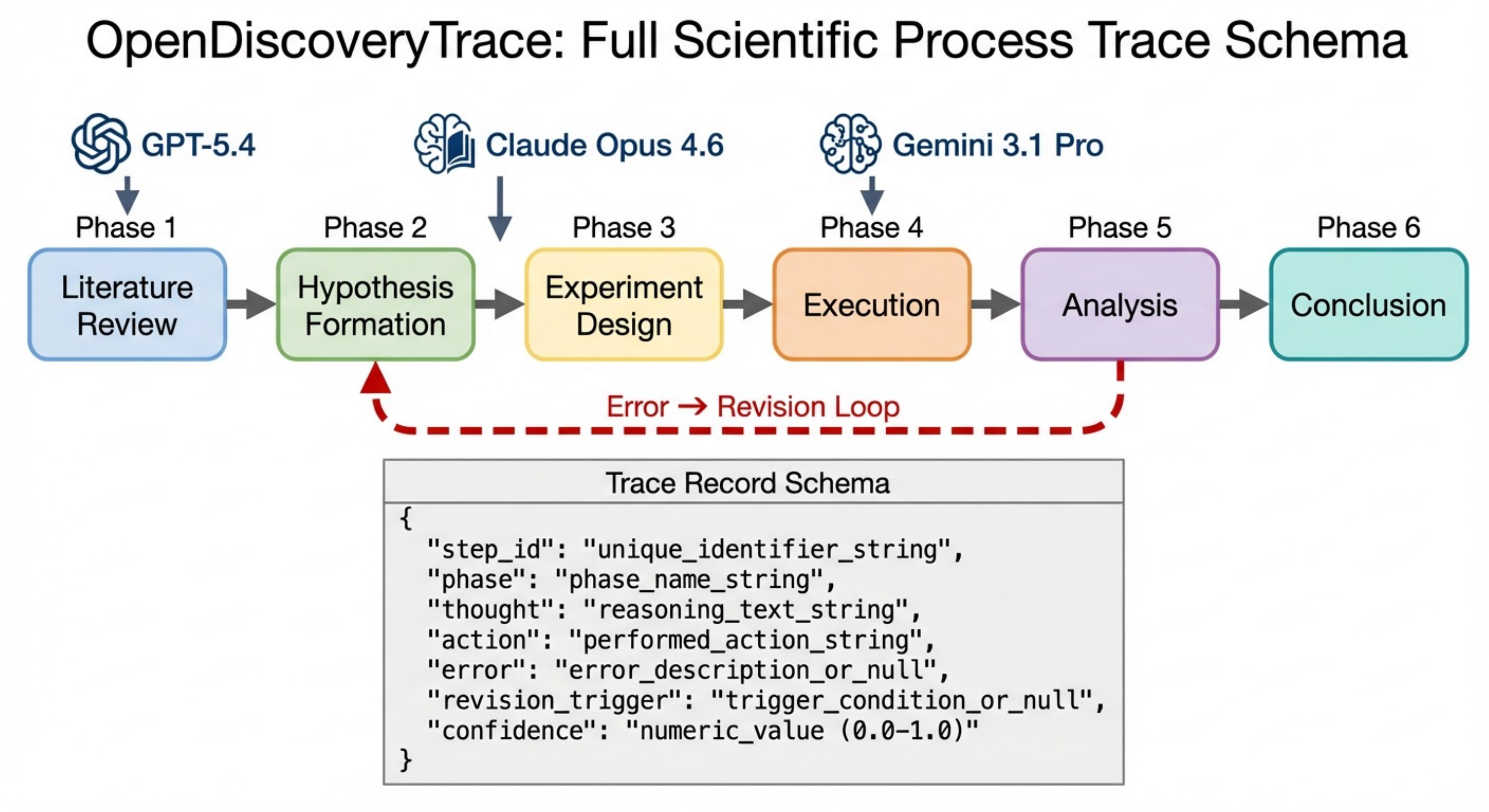}
    \vspace{-1.5em}
    \caption{\footnotesize \dataset: 7 models execute tasks through a six-phase workflow. Each step is a 9-field trace. The 558 trajectories form a corpus of AI scientific process traces.}
    \label{fig:architecture}
    \vspace{-1em}
\end{figure}

\textbf{Why process traces matter.} LLM-as-judge evaluation on 363 frontier trajectories shows all three models achieve comparable success (GPT-5.4: 88.6\%, Gemini: 89.3\%, Claude: 83.9\%), yet Claude Opus 4.6 produces $30{\times}$ more errors than GPT-5.4 ($2.5$ vs.\ $0.08$; $H{=}122$, $p{<}10^{-4}$; Cliff's $\delta{=}0.613$). These errors are qualitatively different: Claude's are 66.7\% tool misuse; GPT-5.4's are 83.6\% reasoning errors. This finding is robust across all four domains in matched-task comparisons (Appendix~\ref{app:matched}). LSTM and Transformer baselines on outcome prediction fail to exceed majority baseline (AUROC${\leq}$0.42), confirming the task is non-trivial (Appendix~\ref{app:baselines}). Output-only benchmarks miss process-level differences entirely.

\vspace{-0.3em}
\section{Acquisition Roadmap}
\vspace{-0.3em}

\textbf{Phase~1 (current).} 558 trajectories across 7 models (3 frontier + 4 open-weight), 124 tasks, 4 domains, plus 60 live-retrieval variants and 6 tool-scaffolded open-weight trajectories. Total cost: {\raise.17ex\hbox{$\scriptstyle\sim$}}\$340 (API + GPU). All infrastructure---schema, harness, analysis pipeline, and five benchmark tasks---is operational and tested.

\textbf{Phase~2 (months 1--2).} Complete all 200 designed tasks across all 7 existing models plus 2 additional open-weight models (Llama~3.1-8B, Gemma-2-9B). Target: 1,800+ trajectories across 9 models, enabling robust cross-architecture comparison at scale with sufficient statistical power for domain-stratified and difficulty-stratified analyses.

\textbf{Phase~3 (months 2--3).} Human expert annotation on a 120-trajectory stratified sample. Three domain-expert annotators independently score five axes: correctness, reasoning quality, tool use efficiency, autonomy level (L1--L4), and error step localization. Target: Krippendorff's $\alpha \geq 0.67$. An LLM-based annotation pilot already achieves $\alpha{=}0.63$--$0.82$ across axes (Appendix~\ref{app:iaa}), suggesting human agreement is attainable.

\textbf{Phase~4 (months 3--6).} Live retrieval integration for all literature-domain tasks via PubMed and PubChem APIs (a 30-task pilot is already included). ELN integration pilot for physical-lab trace capture~\citep{dirnagl2016eln,abolhasani2023selfdriving}, extending the dataset from computational-only to hybrid computational-physical settings.

\textbf{Sustainability.} Semver-versioned schema (v1.0 = current release). Dataset hosted on the Hugging Face Hub, with the harness and analysis code on GitHub (Appendix~\ref{app:access}). Open-source harness enabling community-contributed trajectories. Annual refresh cycles incorporating new frontier and open-weight models as they become available.

\vspace{-0.3em}
\section{Metadata, Governance \& Safety}
\vspace{-0.3em}

\textbf{License:} CC-BY-4.0. \textbf{Format:} Self-contained JSON per trajectory with per-step and aggregate metadata. Datasheet per~\citet{gebru2021datasheets}. \textbf{Release tiers:} (1)~full traces where provider ToS permits; (2)~abstracted summaries (phase transitions, tool calls, errors---without verbatim reasoning) preserving ${>}90\%$ analytical utility. \textbf{Privacy:} No PII; public scientific knowledge only. \textbf{Splits:} 80/20 train/test, domain-stratified, no leakage~\citep{kapoor2023leakage}. Evaluation follows HELM~\citep{liang2023helm}.

\vspace{-0.3em}
\section{Acceleration Potential}
\vspace{-0.3em}

\dataset enables six new capabilities: (1)~\emph{process-aware evaluation} distinguishing reasoning from luck~\citep{wang2023scientific,gil2014amplify}; (2)~\emph{early failure intervention}---68.1\% of first errors at step~0 suggests monitoring initial tool calls could prevent most failures; (3)~\emph{error-type-specific improvement}---Claude needs better tool-use training, GPT-5.4 needs reasoning-level fixes; (4)~\emph{process-aware reward modeling} for RL that credits methodology, not just outcomes; (5)~\emph{trustworthy auditing} via complete decision traces for governance~\citep{king2009automation,kitano2021nobel}; (6)~\emph{reproducible open-weight benchmarking}---120 trajectories from 4 models, no API keys required.

\clearpage
\bibliography{references}
\bibliographystyle{icml2026}

\clearpage
\onecolumn
\appendix
\section*{Appendix}
\setcounter{section}{0}
\renewcommand{\thesection}{\Alph{section}}

\section{Complete Trace Schema}
\label{app:schema}

Every trajectory in \dataset is stored as a self-contained JSON file.
The schema below defines every field. Fields marked \texttt{*} are present in every trajectory; others may be \texttt{null} for certain steps.

\begin{lstlisting}[style=json, caption={Complete \dataset JSON schema for a single trajectory.}, label=lst:schema]
{
  "trajectory_id": "string*  (unique: {task_id}_{model})",
  "task_id":       "string*  (e.g. dd_e01, ms_m03, gn_h02, lt_m05)",
  "domain":        "string*  (drug_discovery | materials_science
                              | genomics | literature)",
  "difficulty":    "string*  (easy | medium | hard)",
  "prompt":        "string*  (the scientific task given to the model)",
  "ground_truth":  "string   (expected answer for easy tasks; null for
                              medium/hard open-ended tasks)",
  "model":         "string*  (gpt-5.4 | claude-opus-4.6 |
                              gemini-3.1-pro | qwen2.5-7b |
                              mistral-7b-v0.3 | phi-3.5-mini |
                              qwen2.5-1.5b)",
  "open_weight":   "boolean  (true for open-weight models)",

  "trajectory": [
    {
      "step_id":    "integer* (0-indexed sequential step number)",
      "timestamp":  "string*  (ISO 8601 UTC, e.g. 2026-04-16T14:30:00Z)",
      "phase":      "string*  (literature_review | hypothesis |
                               experiment_design | execution |
                               analysis | revision | conclusion)",
      "thought":    "string*  (model's reasoning at this step)",
      "action": {
        "type":   "string*  (tool_call | reasoning | conclude)",
        "tool":   "string   (python_exec | web_search |
                             pubmed_search | api_call | none)",
        "input":  "string   (argument passed to the tool)",
        "output": "string   (raw output returned by the tool)"
      },
      "observation":      "string  (processed observation from action)",
      "error": {
        "occurred": "boolean* (true if this step had an error)",
        "type":     "string   (tool_error | api_error | timeout |
                               parse_error | null)",
        "message":  "string   (error message text; null if no error)"
      },
      "revision_trigger": "string  (what prompted a strategy change;
                                    null if no revision at this step)",
      "confidence":       "float   (model's self-reported certainty
                                    [0.0, 1.0]; null if not stated)",
      "raw_response":     "string  (first 3000 chars of model output)",
      "wall_time":        "float   (seconds for this step)"
    }
  ],

  "outcome": {
    "success":            "boolean (true/false/null; null = not yet
                                    evaluated against ground truth)",
    "final_claim":        "string  (model's concluding statement)",
    "confidence":         "float   (confidence on the final claim)",
    "verification": {
      "method":  "string  (ground_truth_match | llm_judge | pending)",
      "result":  "string  (correct | incorrect | partial | pending)",
      "score":   "float   (0.0--1.0 verification score)"
    },
    "failure_type":       "string  (from failure taxonomy; null if
                                    no failure)",
    "recovery_attempted": "boolean (true if the model tried to recover
                                    from an error)",
    "recovery_successful":"boolean (true if recovery led to success;
                                    null if no recovery attempted)"
  },

  "metadata": {
    "total_steps":       "integer* (number of steps in trajectory)",
    "total_tokens_est":  "integer  (estimated token count)",
    "total_tool_calls":  "integer* (number of tool invocations)",
    "total_failures":    "integer* (number of steps with errors)",
    "total_revisions":   "integer* (number of strategy revisions)",
    "wall_time_seconds": "float*   (total wall-clock time)",
    "max_steps_reached": "boolean  (true if hit 30-step limit)",
    "model_version":     "string   (exact model ID used)",
    "temperature":       "float    (sampling temperature; 0.0 for
                                    frontier, varies for open-weight)",
    "collection_timestamp":"string (ISO 8601 collection date)"
  }
}
\end{lstlisting}

\section{Benchmark Comparison with Existing Datasets}
\label{app:comparison}

\begin{table*}[t]
\caption{Comparison of \dataset with existing AI science benchmarks. \dataset is the first benchmark to capture full process traces including error logging, revision tracking, and failure analysis alongside final output evaluation. $\checkmark$ = supported, $\times$ = not supported, $\sim$ = partial support.}
\label{tab:comparison}
\centering
\small
\begin{tabular}{lcccccc}
\toprule
\textbf{Benchmark} & \textbf{Process} & \textbf{Error} & \textbf{Revision} & \textbf{Multi-} & \textbf{Science} & \textbf{Multi-} \\
 & \textbf{Traces} & \textbf{Logging} & \textbf{Tracking} & \textbf{Domain} & \textbf{Specific} & \textbf{Model} \\
\midrule
ScienceAgentBench \citep{chen2025scienceagentbench} & $\times$ & $\times$ & $\times$ & $\checkmark$ & $\checkmark$ & $\checkmark$ \\
DiscoveryBench \citep{majumder2024discoverybench} & $\times$ & $\times$ & $\times$ & $\checkmark$ & $\checkmark$ & $\checkmark$ \\
The AI Scientist \citep{lu2024aiscientist} & $\times$ & $\times$ & $\times$ & $\sim$ & $\checkmark$ & $\checkmark$ \\
PaperBench \citep{starace2025paperbench} & $\times$ & $\times$ & $\times$ & $\sim$ & $\checkmark$ & $\checkmark$ \\
ResearchBench \citep{liu2025researchbench} & $\times$ & $\times$ & $\times$ & $\checkmark$ & $\checkmark$ & $\checkmark$ \\
SciCode \citep{tian2024scicode} & $\times$ & $\times$ & $\times$ & $\checkmark$ & $\checkmark$ & $\checkmark$ \\
MLAgentBench \citep{huang2024mlagentbench} & $\sim$ & $\sim$ & $\times$ & $\times$ & $\sim$ & $\checkmark$ \\
MLE-bench \citep{chan2024mlebench} & $\times$ & $\times$ & $\times$ & $\times$ & $\times$ & $\checkmark$ \\
WebArena \citep{zhou2024webarena} & $\checkmark$ & $\sim$ & $\times$ & $\times$ & $\times$ & $\checkmark$ \\
AgentBench \citep{liu2024agentbench} & $\checkmark$ & $\sim$ & $\times$ & $\times$ & $\times$ & $\checkmark$ \\
\midrule
\textbf{\dataset (ours)} & $\checkmark$ & $\checkmark$ & $\checkmark$ & $\checkmark$ & $\checkmark$ & $\checkmark$ \\
\bottomrule
\end{tabular}
\end{table*}

Table~\ref{tab:comparison} compares \dataset with existing AI science benchmarks and agent trajectory datasets. The key distinction is that \dataset is the only resource providing structured process traces with error logging, revision tracking, and multi-phase scientific workflow annotation alongside final output evaluation. Most existing benchmarks evaluate final artifacts only; agent trajectory datasets like ToolBench and MINT record traces but exclusively for web or software domains, not science.

\section{Dataset Summary Statistics}
\label{app:stats}

\begin{table}[h]
\caption{Complete dataset statistics.}
\label{tab:stats}
\centering\small
\begin{tabular}{lc}
\toprule
\textbf{Metric} & \textbf{Value} \\
\midrule
\textbf{Scale} \\
\quad Total trajectories & 558 \\
\quad Frontier model trajectories & 372 (124 $\times$ 3 models) \\
\quad Open-weight model trajectories & 120 (30 $\times$ 4 models) \\
\quad Live-retrieval variant trajectories & 60 \\
\quad Models represented & 7 (3 frontier + 4 open-weight) \\
\quad Domains covered & 4 \\
\quad Tasks executed per frontier model & 124 \\
\quad Tasks in full bank (designed) & 200 \\
\quad Tasks per domain per frontier model & 31 \\
\midrule
\textbf{Trajectory characteristics (frontier models)} \\
\quad Conclusion rate & 99.7\% \\
\quad Success rate (LLM-judged, $n{=}363$) & 87.3\% \\
\quad Mean steps per trajectory & 4.1 \\
\quad Median steps per trajectory & 2.0 \\
\quad Mean tool calls per trajectory & 3.0 \\
\quad Mean errors per trajectory & 1.0 \\
\quad Mean revisions per trajectory & 0.87 \\
\quad Mean wall time (seconds) & 99.1 \\
\quad Trajectories with $\geq 1$ error & 31.2\% \\
\quad Trajectories with $\geq 1$ revision & 34.4\% \\
\midrule
\textbf{Token and cost estimates} \\
\quad Estimated total tokens (frontier) & ${\sim}$701K \\
\quad Mean response chars per trajectory (Claude) & 11,237 \\
\quad Mean response chars per trajectory (GPT-5.4) & 5,720 \\
\quad Mean response chars per trajectory (Gemini) & 5,663 \\
\quad API cost (frontier models) & ${\sim}$\$300 \\
\quad GPU cost (open-weight models, Modal A10G) & ${\sim}$\$40 \\
\bottomrule
\end{tabular}
\end{table}

\section{Inter-Model Comparison}
\label{app:models}

\begin{table*}[h]
\caption{Complete inter-model comparison for frontier models ($n{=}124$ per model). All tests are Kruskal-Wallis (equivalent to Mann-Whitney $U$ for pairwise). Cliff's $\delta$ effect sizes: negligible ($<$0.147), small (0.147--0.33), medium (0.33--0.474), large ($>$0.474). Values: mean $\pm$ SD. $^{***}$~$p{<}0.001$, $^{**}$~$p{<}0.01$, $^{*}$~$p{<}0.05$, n.s.~not significant.}
\label{tab:model_detailed}
\centering\small
\begin{tabular}{lccccc}
\toprule
\textbf{Metric} & \textbf{GPT-5.4} & \textbf{Claude Opus 4.6} & \textbf{Gemini 3.1 Pro} & \textbf{Significance} & \textbf{Cliff's $\delta$ (max pair)} \\
\midrule
Total steps & 3.5 $\pm$ 3.5 & 4.1 $\pm$ 4.0 & 4.8 $\pm$ 5.4 & n.s. ($p{=}0.158$) & 0.160 (small) \\
Tool calls & 2.5 $\pm$ 3.5 & 2.8 $\pm$ 3.2 & 3.8 $\pm$ 5.2 & n.s. ($p{=}0.214$) & 0.149 (small) \\
Errors & 0.08 $\pm$ 0.3 & 2.5 $\pm$ 3.4 & 0.4 $\pm$ 0.8 & $^{***}$ ($H{=}122$, $p{<}10^{-4}$) & 0.613 (large) \\
Revisions & 0.4 $\pm$ 0.8 & 1.2 $\pm$ 2.0 & 1.0 $\pm$ 2.5 & $^{***}$ ($H{=}15.6$, $p{=}0.0004$) & 0.244 (small) \\
Wall time (s) & 39.6 $\pm$ 37.4 & 190.6 $\pm$ 148.3 & 67.0 $\pm$ 89.9 & $^{***}$ ($p{<}10^{-4}$) & --- \\
Response chars & 5,720 & 11,237 & 5,663 & --- & --- \\
Success rate (judged) & 88.6\% & 83.9\% & 89.3\% & n.s. ($p{=}0.38$) & --- \\
Conclusion rate & 100\% & 99.2\% & 100\% & n.s. ($p{=}0.368$) & --- \\
\bottomrule
\end{tabular}
\end{table*}

The central finding is that \textbf{output metrics (success rate, conclusion rate) show no significant differences}, while \textbf{process metrics (errors, revisions, wall time) show highly significant differences} with large effect sizes. An output-only benchmark would rate these three models as interchangeable; our process traces reveal fundamentally different error-handling behaviors.

\section{Failure Taxonomy}
\label{app:failure}

\textbf{Note on error counts.} Two related but distinct error metrics appear throughout this paper: (i)~\texttt{total\_failures} in trajectory metadata counts \emph{steps where the harness detected an explicit error} (e.g., tool returned an exception, API timeout), and (ii)~the failure taxonomy below counts \emph{all error events}, including both explicit tool errors and keyword-detected reasoning errors (revision triggers such as ``mistake,'' ``let me correct''). The taxonomy totals are therefore higher than the per-step failure counts. Claude averages 2.5 explicit step-errors per trajectory ($124 \times 2.5 = 310$ step-errors), but the taxonomy additionally captures 152 keyword-detected reasoning errors, yielding 462 total error events.

We categorize error events into three types:

\begin{itemize}
\item \textbf{Tool misuse}: Incorrect API call, wrong parameters, malformed input to a tool, or misinterpretation of tool output. Example: passing a gene name to PubChem (which expects a chemical compound name) or calling \texttt{python\_exec} with syntactically invalid code.
\item \textbf{Reasoning error}: Logical fallacy, incorrect inference, flawed calculation, or unsupported conclusion. Detected via keyword-based revision triggers (``mistake'', ``reconsider'', ``let me correct'').
\item \textbf{Hallucination}: Agent fabricates data, citations, or tool outputs. Rare in our data (1 instance across 688 total errors).
\end{itemize}

\begin{table}[h]
\caption{Failure taxonomy breakdown by model.}
\label{tab:failure_taxonomy}
\centering\small
\begin{tabular}{lccc}
\toprule
\textbf{Model} & \textbf{Total Errors} & \textbf{Tool Misuse} & \textbf{Reasoning Error} \\
\midrule
Claude Opus 4.6 & 462 & 308 (66.7\%) & 153 (33.1\%) \\
Gemini 3.1 Pro & 165 & 46 (27.9\%) & 119 (72.1\%) \\
GPT-5.4 & 61 & 10 (16.4\%) & 51 (83.6\%) \\
\midrule
\textbf{Total} & \textbf{688} & \textbf{364 (52.9\%)} & \textbf{323 (46.9\%)} \\
\bottomrule
\end{tabular}
\end{table}

\textbf{Key insight}: Models differ not only in \emph{how many} errors they make but in \emph{what kind}. Claude's errors are predominantly tool misuse --- it attempts more complex tool interactions and fails more often at the tool interface. GPT-5.4 makes far fewer errors overall, and when it does err, the failures are reasoning-level rather than tool-level. This distinction has direct implications for model improvement: Claude would benefit most from better tool-use training, while GPT-5.4's rarer errors require reasoning-level interventions.

\textbf{Limitation}: Error classification uses automated heuristics (keyword matching on error messages and revision triggers). This may undercount implicit reasoning errors that do not trigger explicit revision keywords. Future work will validate against human-labeled error types on the annotated subset.

\section{Matched-Task Within-Domain Comparisons}
\label{app:matched}

To control for task and domain effects, we compare models only on shared tasks within each domain (31 tasks per domain, all three frontier models execute every task).

\begin{table*}[h]
\caption{Matched-task within-domain comparisons ($n{=}31$ shared tasks per domain per model). All tests Kruskal-Wallis. $^{***}$~$p{<}0.001$, n.s.~not significant.}
\label{tab:matched}
\centering\small
\begin{tabular}{llccccc}
\toprule
\textbf{Domain} & \textbf{Metric} & \textbf{Claude Opus 4.6} & \textbf{Gemini 3.1 Pro} & \textbf{GPT-5.4} & \textbf{$H$} & \textbf{Sig.} \\
\midrule
\multirow{2}{*}{Drug Discovery} & Steps & 4.1 $\pm$ 3.1 & 4.7 $\pm$ 5.6 & 4.2 $\pm$ 3.9 & 0.60 & n.s. \\
& Errors & 2.5 $\pm$ 2.8 & 0.4 $\pm$ 0.9 & 0.03 $\pm$ 0.2 & 29.5 & $^{***}$ \\
\midrule
\multirow{2}{*}{Genomics} & Steps & 5.0 $\pm$ 5.8 & 5.9 $\pm$ 5.9 & 3.2 $\pm$ 3.4 & 2.70 & n.s. \\
& Errors & 3.3 $\pm$ 5.6 & 0.5 $\pm$ 1.0 & 0.0 $\pm$ 0.0 & 29.3 & $^{***}$ \\
\midrule
\multirow{2}{*}{Literature} & Steps & 3.9 $\pm$ 4.0 & 4.3 $\pm$ 5.9 & 3.0 $\pm$ 3.5 & 3.80 & n.s. \\
& Errors & 2.0 $\pm$ 2.3 & 0.3 $\pm$ 1.0 & 0.1 $\pm$ 0.3 & 31.8 & $^{***}$ \\
\midrule
\multirow{2}{*}{Materials Science} & Steps & 3.3 $\pm$ 1.8 & 4.5 $\pm$ 3.9 & 3.6 $\pm$ 3.5 & 0.44 & n.s. \\
& Errors & 2.1 $\pm$ 1.7 & 0.3 $\pm$ 0.5 & 0.2 $\pm$ 0.5 & 34.0 & $^{***}$ \\
\bottomrule
\end{tabular}
\end{table*}

The error rate divergence is \textbf{significant in every domain} ($p{<}0.0001$), while step counts are \textbf{non-significant in every domain}. This confirms that the error-handling difference is robust to domain effects and not an artifact of task selection or difficulty confounds.

\section{Difficulty $\times$ Model Interaction}
\label{app:difficulty}

\begin{table}[h]
\caption{Per-difficulty, per-model mean steps and errors.}
\label{tab:difficulty}
\centering\small
\begin{tabular}{llcc}
\toprule
\textbf{Difficulty} & \textbf{Model} & \textbf{Mean Steps} & \textbf{Mean Errors} \\
\midrule
\multirow{3}{*}{Easy ($n{=}147$)} & Claude Opus 4.6 & 1.9 & 0.27 \\
& Gemini 3.1 Pro & 3.4 & 0.27 \\
& GPT-5.4 & 2.0 & 0.00 \\
\midrule
\multirow{3}{*}{Medium ($n{=}153$)} & Claude Opus 4.6 & 5.8 & 4.14 \\
& Gemini 3.1 Pro & 5.0 & 0.45 \\
& GPT-5.4 & 4.9 & 0.12 \\
\midrule
\multirow{3}{*}{Hard ($n{=}72$)} & Claude Opus 4.6 & 4.9 & 3.54 \\
& Gemini 3.1 Pro & 7.4 & 0.42 \\
& GPT-5.4 & 3.4 & 0.17 \\
\bottomrule
\end{tabular}
\end{table}

The error-handling divergence is concentrated in \textbf{medium and hard tasks}. Easy tasks produce uniformly short, low-error trajectories across all models (Claude averages just 0.27 errors on easy tasks vs.\ 4.14 on medium). This validates both the difficulty stratification and the intuition that process differences emerge most strongly on challenging scientific problems.

\section{Benchmark Task Baselines}
\label{app:baselines}

\subsection{Task 1: Trajectory Outcome Prediction}

\textbf{Setup}: Given all step-level features from a complete trajectory, predict binary success. With the expanded LLM-judged evaluation ($n{=}363$), we use this larger set for training and testing. Features: total steps, tool calls, errors, revisions, unique tools, unique phases, response length, error type count.

\begin{table}[h]
\caption{Outcome prediction baselines (5-fold stratified CV).}
\label{tab:baselines}
\centering\small
\begin{tabular}{lcc}
\toprule
\textbf{Method} & \textbf{Accuracy} & \textbf{AUROC} \\
\midrule
Majority baseline & 0.693 & 0.500 \\
Logistic Regression & 0.685 $\pm$ 0.046 & 0.496 $\pm$ 0.096 \\
Random Forest & 0.614 $\pm$ 0.049 & 0.452 $\pm$ 0.091 \\
Gradient Boosting & 0.614 $\pm$ 0.033 & 0.426 $\pm$ 0.066 \\
LSTM (2-layer, $h{=}64$) & 0.693 $\pm$ 0.015 & 0.422 $\pm$ 0.124 \\
Transformer (2-layer, 4-head) & 0.693 $\pm$ 0.015 & 0.369 $\pm$ 0.093 \\
\bottomrule
\end{tabular}
\end{table}

\textbf{Finding}: No method exceeds the majority baseline accuracy (0.693). AUROC values are near or below chance. This confirms that \textbf{outcome prediction from trajectory-level features is genuinely difficult}: neither aggregate statistics (LR, RF, GBT) nor sequential patterns (LSTM, Transformer) on the current 8-dimensional feature set are predictive of success. This motivates future work on richer representations --- e.g., semantic embeddings of thought content or graph-structured encodings of tool-call dependencies.

\subsection{Task 2: Error Localization}

\textbf{Setup}: Among the 116 trajectories containing at least one error, identify the first step where reasoning went wrong.

\textbf{Result}: 68.1\% of first errors occur at step 0 (the first tool invocation). The ``always predict step 0'' heuristic achieves 68.1\% accuracy; ``always predict last step'' achieves 0\%. Random baseline: ${\sim}$7\% (1/mean-steps).

\textbf{Caveat}: The step-0 concentration partially reflects harness design (the first substantive action is typically a tool call). Future work should vary tool-initialization ordering to disambiguate harness artifact from generalizable agent behavior.

\subsection{Task 3: Claim Verification}

Success rates with Wilson score 95\% confidence intervals, evaluated via cross-model LLM-as-judge on the full frontier set ($n{=}363$):
\begin{itemize}
\item GPT-5.4: 88.6\% [81.8\%, 93.1\%] ($n{=}123$)
\item Claude Opus 4.6: 83.9\% [76.2\%, 89.4\%] ($n{=}118$)
\item Gemini 3.1 Pro: 89.3\% [82.6\%, 93.7\%] ($n{=}122$)
\end{itemize}

The expanded evaluation (from $n{=}127$ ground-truth-only to $n{=}363$ via LLM judging) tightens confidence intervals substantially and includes medium and hard tasks. The 9 trajectories not judged (3.6\%) produced malformed claims that the judge could not parse. Human expert evaluation on a 120-trajectory sample remains a Phase~3 priority to anchor absolute accuracy.

\subsection{Task 5: Process Quality Scoring}

We define a composite process quality score as the unweighted mean of four normalized [0,1] components: \emph{efficiency} ($1 - \text{steps}/\text{max\_steps}$), \emph{tool diversity} (unique tools / max unique tools), \emph{low errors} ($1 - \text{errors}/\text{max\_errors}$), and \emph{conclusion reached} (binary). Equal weights are used as a deliberate baseline; optimizing weights via correlation with human quality judgments is a Phase~3 objective.
\begin{itemize}
\item GPT-5.4: 0.747 $\pm$ 0.008
\item Gemini 3.1 Pro: 0.747 $\pm$ 0.011
\item Claude Opus 4.6: 0.724 $\pm$ 0.067
\end{itemize}
Kruskal-Wallis $H{=}149.9$, $p{<}0.0001$. Claude scores significantly lower due to its higher error rate, despite matching success rates. This demonstrates that \textbf{process quality scoring reveals quality differences invisible to binary success metrics}.

\section{LLM-Based Inter-Annotator Agreement}
\label{app:iaa}

We annotated a 60-trajectory stratified sample (5 per model per domain) using GPT-5.4-mini and Claude Sonnet 4.6 as annotators. Each annotator rated every trajectory on four axes (1--5 ordinal scale for the first three; L1--L4 categorical for autonomy).

\begin{table}[h]
\caption{LLM-based inter-annotator agreement (Krippendorff's $\alpha$).}
\label{tab:iaa}
\centering\small
\begin{tabular}{lcc}
\toprule
\textbf{Axis} & \textbf{$\alpha$} & \textbf{Interpretation} \\
\midrule
Correctness & 0.629 & Moderate \\
Reasoning quality & 0.711 & Substantial \\
Tool use efficiency & 0.717 & Substantial \\
Autonomy level (L1--L4) & 0.820 & Near-perfect \\
\bottomrule
\end{tabular}
\end{table}

The autonomy taxonomy (L1--L4) exceeds the $\alpha {\geq} 0.67$ target, providing initial evidence that the proposed levels are reliably distinguishable. Weighted agreement (within 1 point on 5-point scales) exceeds 0.85 across all axes. We note that LLM-based annotation is a proxy for human annotation; the planned Phase~3 human expert study (3 annotators, 120 trajectories) will provide definitive IAA measurements.

\section{Tool Usage and Latency Statistics}
\label{app:tools}

\begin{table}[h]
\caption{Tool invocation counts by model (across 124 frontier trajectories each).}
\label{tab:tools}
\centering\small
\begin{tabular}{lccc}
\toprule
\textbf{Tool} & \textbf{Claude} & \textbf{Gemini} & \textbf{GPT-5.4} \\
\midrule
\texttt{python\_exec} & 341 & 372 & 163 \\
\texttt{web\_search} & 7 & 49 & 72 \\
\texttt{pubmed\_search} & 1 & 41 & 35 \\
\texttt{api\_call} & 1 & 3 & 31 \\
\midrule
\textbf{Total} & 350 & 465 & 301 \\
\bottomrule
\end{tabular}
\end{table}

Python execution dominates (78.5\% of 1,116 total calls). GPT-5.4 uses a notably more diverse tool mix (23.9\% web search, 11.6\% PubMed, 10.3\% API calls) compared to Claude (98\% Python). Mean per-tool latencies: \texttt{python\_exec} 27.5s, \texttt{pubmed\_search} 11.2s, \texttt{web\_search} 8.8s, \texttt{api\_call} 6.8s.

\textbf{Note on web search}: The current release uses simulated web search (models use parametric knowledge) for frontier trajectories. We include 60 live-retrieval variant trajectories using real PubMed and PubChem API calls to assess ecological validity. Phase~4 will expand live retrieval to all literature-domain tasks.

\section{Open-Weight Model Details}
\label{app:openweight}

To enable fully reproducible baselines without proprietary API access, we generated 120 trajectories from four open-weight models, each executing the same 30 tasks (first 30 from the task bank: drug discovery easy and medium).

\begin{table}[h]
\caption{Open-weight model specifications and performance.}
\label{tab:openweight}
\centering\small
\begin{tabular}{lccc}
\toprule
\textbf{Model} & \textbf{Parameters} & \textbf{GPU} & \textbf{Mean time/task} \\
\midrule
Qwen2.5-7B-Instruct & 7B & A10G & ${\sim}$15s \\
Mistral-7B-Instruct-v0.3 & 7B & A10G & ${\sim}$15s \\
Phi-3.5-mini-instruct & 3.8B & A10G & ${<}$1s \\
Qwen2.5-1.5B-Instruct & 1.5B & V100 & ${\sim}$7s \\
\bottomrule
\end{tabular}
\end{table}

The 120 single-response trajectories demonstrate that the schema accommodates any model interface. Additionally, we ran Qwen2.5-7B through the \textbf{full multi-step harness with tool scaffolding} on an A100 GPU, producing 6 trajectories (avg 10.3 steps, 1.7 tool calls, 1.7 errors per trajectory, 100\% conclusion rate). These tool-scaffolded trajectories enable direct comparison with frontier models under identical conditions: Qwen2.5-7B shows a step count comparable to frontier models (10.3 vs.\ 3.5--4.8) but with higher error rates, demonstrating that the harness can surface meaningful process-level differences across the open/frontier divide. Future work will expand tool-scaffolded open-weight coverage to all 30 tasks and additional models.

\section{Task Bank Examples}
\label{app:tasks}

The full task bank comprises 200 tasks (50 per domain). Below are representative examples at each difficulty level. The complete task bank is available in \texttt{task\_bank.json} in the release.

\textbf{Drug Discovery.}
\begin{itemize}
\item \textbf{Easy}: ``Retrieve the molecular weight, LogP, and number of hydrogen bond donors for aspirin. Report the exact values.''
\item \textbf{Medium}: ``Find all known kinase inhibitors approved by the FDA that target VEGFR-2. For each, report the drug name, approval year, and whether it also inhibits other kinases.''
\item \textbf{Hard}: ``Propose a novel drug repurposing hypothesis for an existing FDA-approved drug to treat Alzheimer's disease. Base your hypothesis on molecular target analysis, pathway overlap, and available clinical evidence. Design a computational experiment to validate your hypothesis.''
\end{itemize}

\textbf{Materials Science.}
\begin{itemize}
\item \textbf{Easy}: ``What is the band gap of gallium arsenide (GaAs)? Is it a direct or indirect band gap semiconductor?''
\item \textbf{Medium}: ``Design a high-entropy alloy (HEA) composition for high-temperature applications. Select 5 elements, justify each choice based on Hume-Rothery rules, and predict the likely crystal structure using the VEC criterion.''
\item \textbf{Hard}: ``Propose a novel solid-state electrolyte composition for all-solid-state lithium batteries that balances ionic conductivity, electrochemical stability window, and mechanical properties. Justify using first-principles reasoning and literature evidence.''
\end{itemize}

\textbf{Genomics.}
\begin{itemize}
\item \textbf{Easy}: ``What is the chromosomal location and function of the TP53 gene? How many exons does it have?''
\item \textbf{Medium}: ``Design a CRISPR guide RNA to knock out the human PCSK9 gene. Specify the target exon, the 20-nt guide sequence, the PAM site, and predict off-target effects.''
\item \textbf{Hard}: ``Propose a computational pipeline to identify novel synthetic lethal gene pairs in pancreatic cancer. Design the approach using DepMap, COSMIC, and STRING, specify the statistical framework, and predict at least one novel interaction.''
\end{itemize}

\textbf{Scientific Literature.}
\begin{itemize}
\item \textbf{Easy}: ``Find the original paper describing the transformer architecture (`Attention Is All You Need'). Report the full citation, venue, and citation count.''
\item \textbf{Medium}: ``Identify contradictions in the literature about whether transformer-based protein language models can predict protein function more accurately than alignment-based methods. Find papers supporting each side.''
\item \textbf{Hard}: ``Write a critical analysis of the claim that `AI will replace human scientists within 10 years.' Gather evidence from proponents and skeptics, evaluate argument strength, and formulate an evidence-based position.''
\end{itemize}

\section{Agent Harness Details}
\label{app:harness}

The agent harness is a Python script (\texttt{agent\_harness.py}) that orchestrates trajectory generation. Key design choices:

\begin{itemize}
\item \textbf{System prompt}: A standardized 6-phase scientific workflow (literature review $\to$ hypothesis $\to$ experiment design $\to$ execution $\to$ analysis $\to$ conclusion). All models receive the identical system prompt.
\item \textbf{Tool suite}: Four tools: \texttt{python\_exec} (sandboxed subprocess with 60s timeout), \texttt{web\_search} (simulated), \texttt{pubmed\_search} (real E-utilities API), \texttt{api\_call} (PubChem, UniProt, Materials Project).
\item \textbf{Step limit}: Maximum 30 steps per trajectory. Models are nudged to conclude starting at step 15 and explicitly prompted at step 25.
\item \textbf{Temperature}: 0.0 for all frontier models (deterministic). Open-weight models use \texttt{do\_sample=False}.
\item \textbf{Checkpointing}: Each trajectory is saved as a JSON file immediately upon completion.
\item \textbf{Concurrency}: Up to 4 concurrent API calls per model to respect rate limits.
\item \textbf{Error handling}: API errors (timeouts, rate limits) are logged in the trajectory and the model is given an opportunity to recover.
\end{itemize}

\textbf{Sensitivity to harness design}: The 6-phase prompt, 30-step cap, and tool suite are design choices that affect trajectory characteristics. Models may collapse phases when the prompt encourages comprehensive responses (as observed with Claude's single-response pattern in our earlier pilot). Future work should systematically vary these parameters (step limits of 10/30/60, alternative prompt structures, expanded tool suites) to quantify sensitivity.

\section{Code and Data Availability}
\label{app:access}

Code, data, and the complete task bank are publicly available under CC-BY-4.0:
\begin{itemize}
\item \textbf{Dataset} (all trajectory JSON files, task bank, and analysis results): \url{https://huggingface.co/datasets/aayambansall/OpenDiscoveryTrace}
\item \textbf{Code} (agent harness, analysis pipelines, benchmark baselines, sample trajectories, and paper source): \url{https://github.com/aayambansal/OpenDiscoveryTrace}
\end{itemize}
The GitHub repository contains the agent harness script for trajectory generation and the analysis pipelines that reproduce all figures and statistics; the trajectory JSON files, task bank, and analysis results are hosted on the Hugging Face Hub.

\textbf{Reproducing frontier trajectories} requires API keys for the three frontier model providers. The harness script accepts a \texttt{--model} flag to select any supported model and a \texttt{--max-tasks} flag to control the number of tasks.

\textbf{Reproducing open-weight trajectories} requires only a single GPU (V100 16GB or better) and access to open-weight model weights. No proprietary API keys are needed, enabling fully independent replication.

\textbf{Running analysis}: A single script regenerates all figures, statistical tables, and benchmark baselines from the raw trajectory JSON files, ensuring end-to-end reproducibility of all reported results.

\end{document}